\documentclass[twoside,11pt]{article}

\usepackage[abbrvbib, preprint]{jmlr2e}
\usepackage{amsmath}
\usepackage{algorithm}

\jmlrheading{}{2024}{}{}{}{}{Roumen Nikolaev Popov}

\ShortHeadings{Analytical Approximation of the ELBO Gradient}{Roumen Nikolaev Popov}

\begin{document}

\title{Analytical Approximation of the ELBO Gradient \\ in the Context of the Clutter Problem}

\author{\begin{center}\name Roumen Nikolaev Popov \\
    \small\sc roumenpopov@hotmail.com \\
    \small\rm 16 Apr 2024\end{center}}

\maketitle

\begin{abstract}
We propose an analytical solution for approximating the gradient of the Evidence Lower Bound (ELBO) in variational inference problems where the statistical model is a Bayesian network consisting of observations drawn from a mixture of a Gaussian distribution embedded in unrelated clutter, known as the clutter problem. The method employs the reparameterization trick to move the gradient operator inside the expectation and relies on the assumption that, because the likelihood factorizes over the observed data, the variational distribution is generally more compactly supported than the Gaussian distribution in the likelihood factors. This allows efficient local approximation of the individual likelihood factors, which leads to an analytical solution for the integral defining the gradient expectation. We integrate the proposed gradient approximation as the expectation step in an EM (Expectation Maximization) algorithm for maximizing ELBO and test against classical deterministic approaches in Bayesian inference, such as the Laplace approximation, Expectation Propagation and Mean-Field Variational Inference. The proposed method demonstrates good accuracy and rate of convergence together with linear computational complexity.
\end{abstract}

\section{Introduction}
\label{sec:introduction}

Variational inference provides a viable, deterministic alternative to stochastic sampling methods, such as Markov chain Monte Carlo (MCMC), for approximating the generally intractable marginal likelihood in Bayesian inference problems, as discussed in \cite{Blei2017}. This is achieved by introducing a variational distribution as approximation to the posterior density and then minimizing the Kullback-Leibler (KL) divergence between the two, thus turning the inference problem into an optimization problem over the parameters of the variational distribution (\cite{Jordan1999}, \cite{Bishop2006}, \cite{Zhang2019}). The minimization of the KL divergence is realized  through maximization of a lower bound on the log marginal likelihood, known as the Evidence Lower Bound (ELBO), and the aim is that ELBO might be tractable or easy to approximate where the marginal likelihood is not. This is not always the case, however, and particularly for the clutter problem \cite{Minka2001phd} points out that ELBO is neither tractable nor easy to approximate analytically .

The clutter problem is a toy Bayesian inference problem described in \cite{Minka2001} that has a statistical model defined by a Bayesian network where the observations are generated from a mixture of a Gaussian distribution with known covariance embedded in unrelated clutter. It can also be viewed as a model for measuring a physical quantity where the measurements are corrupted by Gaussian noise and outliers, and in that respect may be useful in embedded applications at the edge or in safety-critical applications such as self-driving cars and industrial processes where undetected outliers in the sensory data have the potential to cause catastrophic failure of the system.

A classical approach for solving the intractability problem of ELBO is to use the mean-field approximation as described in \cite{Bishop2006}, which is based on the assumption that the variational distribution factorizes over the latent variables. However, in the case of the clutter problem the assumption holds poorly, which results in low accuracy of the approximation, demonstrated in \cite{Minka2001}.

Another approach, popularized more recently in \cite{Paisley2012} and \cite{Kingma2014} circumvents the intractability of ELBO by employing stochastic approximation of the ELBO gradient rather than of ELBO itself. This is applicable in cases where computation of the marginal likelihood is not required and we are only interested in maximizing ELBO to optimize the parameters of the variational distribution. Since the definition of ELBO can be regarded as an expectation over the variational distribution, the central premise of this method is to convert the gradient of the expectation into an expectation of a gradient and then stochastically approximate the expectation. In \cite{Paisley2012} the conversion is achieved by using the identity $\nabla_\psi q(\theta|\psi) = q(\theta|\psi)\nabla \ln q(\theta|\psi)$, where $q(\theta|\psi)$ is the variational distribution, $\theta$ comprises the latent variables and $\psi$ represents the parameters of the variational distribution. This approach is known as the log-derivative trick or the score function estimator and while broadly applicable, typically suffers from a high variance of the estimated gradient (\cite{Jankowiak2018}). To avoid the high variance issue, \cite{Kingma2014} adopt a different approach and perform the conversion by employing a reparameterization of the variational distribution in which the latent variable is expressed in terms of a differentiable transformation of an auxiliary random variable with independent distribution. This approach is known as the reparameterization trick or the pathwise gradient estimator and while not as broadly applicable as the score function estimator, generally exhibits lower variance (\cite{Jankowiak2018}).

The stochastic approximation of the ELBO gradient has a primary advantage in that it is broadly applicable and can work with high dimensional latent spaces (\cite{Ranganath2014}), which best benefits application in complex, offline inference problems where the stochastic nature of the method and its usually high computational cost are not an issue. In contrast, real time embedded applications at the edge and safety-critical applications require fast and deterministic inference algorithms. To address this need we propose a method that, in the context of the clutter problem and based on the reparameterization trick, analytically approximates the ELBO gradient and provides a solution for maximizing ELBO by integrating the approximation as the expectation step in an EM (Expectation Maximization) algorithm. The proposed approximation is tested against classical deterministic approaches in Bayesian inference, such as the Laplace approximation, Expectation Propagation and Mean-Field Variational Inference, and demonstrates good accuracy and rate of convergence together with linear computational complexity. The main target application of the method is modelling of 1D sensor readings corrupted by approximately Gaussian noise and outliers within the domain of embedded real time systems and safety critical systems.

\section{Preliminaries}
\label{sec:preliminaries}

\subsection{ELBO Definition}
\label{elbodefinition}

Let $X$ be a set of observations and $Z$ a set of latent variables with a joint distribution $p(X,Z)$. We are interested in approximating the posterior distribution $p(Z|X)$ and the marginal likelihood $p(X)$. For any variational probability distribution $q(Z)$ the log marginal likelihood can be written as follows, as described in \cite{Bishop2006}:
\begin{align}
&\ln p(X) = \mathcal{L}(q(Z)) + \text{KL}(q(Z)||p(Z|X)) \\
&\mathcal{L}(q(Z)) = \int q(Z)\ln\left\{\frac{p(X,Z)}{q(Z)}\right\}dZ \\
&\text{KL}(q(Z)||p(Z|X)) = -\int q(Z)\ln\left\{\frac{p(Z|X)}{q(Z)}\right\}dZ
\end{align}
Since $\text{KL}(q(Z)||p(Z|X))$ is the KL divergence and is always positive, it follows that maximizing $\mathcal{L}(q(Z))$ is equivalent to minimizing $\text{KL}(q(Z)||p(Z|X))$. As such, the quantity $\mathcal{L}(q(Z))$ is a lower bound on the log marginal likelihood and is known as ELBO.

\subsection{Clutter Problem Definition}
\label{subsec:clutterproblem}

The clutter problem is a toy Bayesian inference problem described in \cite{Minka2001} that has a statistical model defined by a Bayesian network where the observations are generated from a mixture of a Gaussian distribution with known covariance embedded in unrelated clutter. In this case, unrelated means that the clutter distribution is independent from the Gaussian distribution in the mixture. The observation density is given by:
\begin{equation} \label{eq:observation}
p(x|\mu) = (1 - w)\mathcal{N}(x;\mu,\Sigma_g) + wP_c(x)
\end{equation}
where $w$ is the probability of having a clutter sample and $P_c(x)$ is the clutter distribution, both of which are known. If we have a set of independent observations $X = \{x_1,...,x_n\}$ and $p(\mu)$ is the prior distribution over $\mu$, then the joint distribution of $X$ and $\mu$ becomes:
\begin{equation}
p(X,\mu) = p(\mu)\prod_{i}p(x_i|\mu)
\end{equation}
Since the joint distribution is a product of sums, we can not really use belief propagation because the belief state for $\mu$ is a mixture of $2^n$ Gaussians (\cite{Minka2001}), which makes the computational complexity prohibitive in cases of more than a few observations.

Applying variational inference to the problem produces the following expression for ELBO, where $q(\mu)$ is the variational distribution:
\begin{equation} \label{eq:elboclutterproblem}
\mathcal{L}(q(\mu)) = \int q(\mu)\sum_{i}\ln\mathcal{L}_i(\mu|x_i)d\mu + \int q(\mu)\ln p(\mu)d\mu - \int q(\mu)\ln q(\mu)d\mu
\end{equation}
and $\mathcal{L}_i(\mu|x_i) =  (1 - w)\mathcal{N}(\mu;x_i,\Sigma_g) + wP_c(x_i)$ are the likelihood factors. Due to the terms $\ln\mathcal{L}_i(\mu|x_i)$, which are logarithms of sums, ELBO is not directly analytically tractable in this case and has to be approximated.

\subsection{The Reparameterization Trick}
\label{subsec:reparameterize}

As described in \cite{Kingma2014}, the reparameterization trick involves expressing the latent variable in terms of a differentiable transformation of an auxiliary random variable with independent distribution, which results in moving the gradient operator inside the expectation. Let $z$ be a continuous latent variable having a distribution $q_\phi(z)$, where $\phi$ are the parameters of the distribution. The latent variable can be expressed as $z = g_\phi(\epsilon)$, where $g_\phi$ is the differentiable transformation and $\epsilon$ is the auxiliary random variable having an independent distribution $q_0(\epsilon)$. Using the reparameterization trick, an expectation over $q_\phi(z)$ can be reformulated as follows:
\begin{equation}
E_q[f(z)] = \int q_\phi(z)f(z)dz = \int q_0(\epsilon)f(g_\phi(\epsilon))d\epsilon
\end{equation}
For the case of a univariate normal distribution $z \sim q_\phi(z) = \mathcal{N}(\mu,\sigma^2)$ a valid reparameterization is $z = \mu + \sigma\epsilon$ (\cite{Kingma2014}), where $\epsilon$ is the auxiliary random variable having a standard normal distribution $\epsilon \sim q_0(\epsilon) = \mathcal{N}(0,1)$.

\section{Analytical Approximation of the ELBO Gradient}
\label{sec:approximation}

We develop the proposed method in the context of the clutter problem for one dimensional observation data. In this case ELBO takes the form as specified in equation (\ref{eq:elboclutterproblem}), where $x_i$ are one dimensional and the Gaussian distribution in the likelihood factors is therefore univariate with variance $v_g$. We use the reparameterization trick to move the gradient operator inside the expectation, with the goal of eliminating the logarithms that are otherwise difficult to directly approximate analytically. The method then relies on the assumption that, because the likelihood factorizes over the observed data, the variational distribution is generally more compactly supported than the Gaussian distribution $\mathcal{N}(\mu;x_i,v_g)$ in the likelihood factors. We use the more casual meaning of the term "support" to refer to the region where the function is meaningfully different from zero and in the case of Gaussian functions a more compact support is equivalent to smaller variance. The smaller support region of the variational distribution allows efficient local approximation of the individual likelihood factors with exponentiated quadratics by employing Taylor series expansion because the error of the local approximation remains relatively low within the small support region. As a result of the approximation with exponentiated quadratics the integral defining the gradient expectation becomes tractable and can be solved analytically. It is important to note that while the approximation of each likelihood factor is local, the resulting approximation of the gradient is not because the local point for each likelihood factor is different. We use the proposed gradient approximation to maximize ELBO by integrating it in the expectation step of an EM algorithm (\cite{Dempster1977}), which is described in section \ref{sec:maxelbo}. The rest of this section is organized into three subsections that develop the main steps of the proposed solution (applying the reparameterization trick, locally approximating the likelihood factors and analytically solving the expectation integrals), a subsection where we briefly discuss possible extension to multidimensional data and a subsection where we examine the applicability to non-Gaussian distributions. We start by restricting the variational distribution $q(\mu)$ to the family of normal distributions $\mathcal{N}(\mu_q,v_q)$ and assume a conjugate prior $p(\mu) = \mathcal{N}(\mu_p,v_p)$.

\subsection{Applying the Reparameterization Trick}
\label{subsec:applyreparameterize}

Using the reparameterization trick for the case of a univariate normal distribition we have $\mu = \mu_q + \sqrt{v_q}\epsilon$ and the ELBO gradient can be expressed as follows:
\begin{equation} \label{eq:elbogradient0}
\nabla\mathcal{L}(q(\mu)) = \int q_0(\epsilon)\sum_{i}\nabla\ln\mathcal{L}_i(\epsilon|(x_i - \mu_q)/\sqrt{v_q})d\epsilon + \nabla\int q(\mu)\ln p(\mu)d\mu - \nabla\int q(\mu)\ln q(\mu)d\mu
\end{equation}
where the gradient operator is over the parameters of the variational distribution $\mu_q$ and $v_q$, and $\mathcal{L}_i(\epsilon|(x_i - \mu_q)/\sqrt{v_q}) = (1 - w)\mathcal{N}(\epsilon;(x_i - \mu_q)/\sqrt{v_q},v_g/v_q)/\sqrt{v_q} + wP_c(x_i)$ are the reparameterized likelihood factors. Applying the gradient operator to the logarithms and carrying out the integration in the tractable terms the expression for the ELBO gradient becomes:
\begin{equation} \label{eq:elbogradient1}
\nabla\mathcal{L}(q(\mu)) = \int q_0(\epsilon)\sum_{i}\frac{\nabla\mathcal{L}_i(\epsilon|(x_i - \mu_q)/\sqrt{v_q})}{\mathcal{L}_i(\epsilon|(x_i - \mu_q)/\sqrt{v_q})}d\epsilon - \frac{1}{2}\nabla\frac{(\mu_p - \mu_q)^2}{v_p} + \frac{1}{2}\nabla\ln\frac{v_q}{v_p} - \frac{1}{2}\nabla\frac{v_q}{v_p}
\end{equation}
Executing the gradient operator over $\mu_q$ and taking the sum outside the integral we obtain:
\begin{equation} \label{eq:elbogradu}
G_{\mu_q} = \frac{\partial \mathcal{L}(q(\mu))}{\partial \mu_q} = -\sqrt{v_q}\sum_{i}\int q_0(\epsilon)\pi_i(\epsilon)\frac{\epsilon - (x_i - \mu_q)/\sqrt{v_q}}{v_g}d\epsilon + \frac{(\mu_p - \mu_q)}{v_p}
\end{equation}
where
\begin{equation} \label{eq:pi_i}
\pi_i(\epsilon) = \frac{(1 - w)\mathcal{N}(\epsilon;(x_i - \mu_q)/\sqrt{v_q},v_g/v_q)/\sqrt{v_q}}{(1 - w)\mathcal{N}(\epsilon;(x_i - \mu_q)/\sqrt{v_q},v_g/v_q)/\sqrt{v_q} + wP_c(x_i)}
\end{equation}
Performing the same for the gradient over $v_q$ results in the expression:
\begin{equation} \label{eq:elbogradv}
G_{v_q} = \frac{\partial \mathcal{L}(q(\mu))}{\partial v_q} = -\frac{1}{2}\left[\sum_{i}\int q_0(\epsilon)\pi_i(\epsilon)\frac{\epsilon(\epsilon - (x_i - \mu_q)/\sqrt{v_q})}{v_g}d\epsilon - \frac{1}{v_q} + \frac{1}{v_p}\right]
\end{equation}

\subsection{Local Approximation of the Likelihood Factors}
\label{subsec:approximatefactors}

\begin{figure}[t!]
\includegraphics[width=1.0\linewidth]{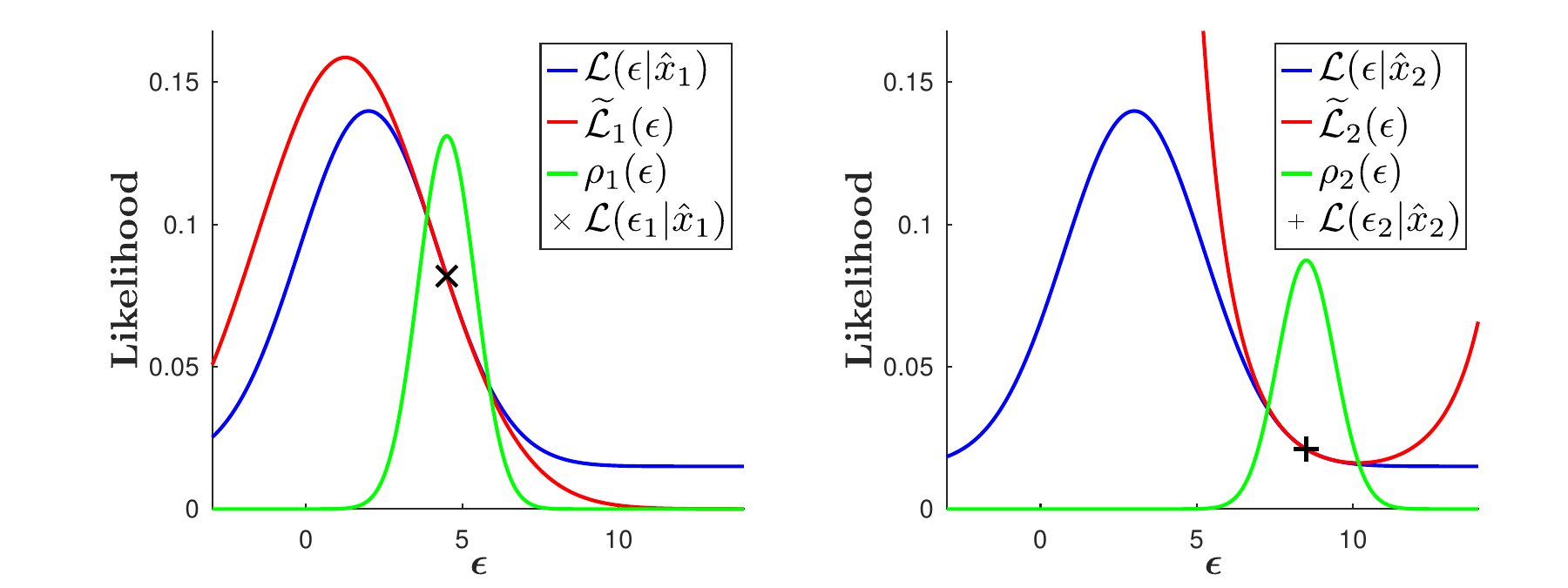}
\caption{Local approximation of the reparameterized likelihood factors}
\label{fig:fig1}
\end{figure}

To construct an analytical expression for the ELBO gradient we need to be able to analytically approximate the integrals in equation (\ref{eq:elbogradu}) and (\ref{eq:elbogradv}). If we can efficiently approximate the term $q_0(\epsilon)\pi_i(\epsilon)$ with a Gaussian then both integrals become tractable. Conveniently, $q_0(\epsilon)$ and the numerator of $\pi_i(\epsilon)$ are both Gaussian and therefore their product is also Gaussian, which we denote $\rho_i(\epsilon)$:
\begin{equation}
\rho_i(\epsilon) = (1 - w)\frac{1}{\sqrt{2(v_g + v_q)\pi}}\exp\left(-\frac{1}{2}\frac{(x_i - \mu_q)^2}{v_g + v_q}\right)\mathcal{N}(\epsilon;\epsilon_i,v_i)
\end{equation}
where $\epsilon_i = \sqrt{v_q}(x_i - \mu_q)/(v_g + v_q)$ and $v_i = v_g/(v_g + v_q)$. Since we have assumed that $q(\mu)$ is more compactly supported than $\mathcal{N}(\mu;x_i,v_g)$, it follows that $q_0(\epsilon)$ is also more compactly supported than $\mathcal{N}(\epsilon;(x_i - \mu_q)/\sqrt{v_q},v_g/v_q)$ and therefore $\rho_i(\epsilon)$ is more compactly supported than $\mathcal{N}(\epsilon;(x_i - \mu_q)/\sqrt{v_q},v_g/v_q)$ as well. Consequently, the reparameterized likelihood factors $\mathcal{L}_i(\epsilon|(x_i - \mu_q)/\sqrt{v_q})$, which are the denominators in $\pi_i(\epsilon)$, can be efficiently approximated locally with exponentiated quadratics as illustrated in figure \ref{fig:fig1}. The quadratics are determined by a second order Taylor series expansion of the logarithm of the factors calculated at the point $\epsilon_i$ and the approximation is specified by the following expressions:
\begin{align}
&\mathcal{L}_i(\epsilon) \approx \mathcal{\widetilde L}_i(\epsilon) = \exp\left(\frac{1}{2}\ln(\mathcal{L}_i(\epsilon))''|_{\epsilon = \epsilon_i}(\epsilon - \epsilon_i)^2 + \ln(\mathcal{L}_i(\epsilon))'|_{\epsilon = \epsilon_i}(\epsilon - \epsilon_i) + \ln\mathcal{L}_i(\epsilon_i)\right) \\
&\ln(\mathcal{L}_i(\epsilon))''|_{\epsilon = \epsilon_i} = \pi_i(\epsilon_i)\left(\left(1 - \pi_i(\epsilon_i)\right)\left(\epsilon_i - (x_i - \mu_q)/\sqrt{v_q}\right)^2v_q/v_g - 1\right)v_q/v_g \\
&\ln(\mathcal{L}_i(\epsilon))'|_{\epsilon = \epsilon_i} = -\pi_i(\epsilon_i)\left(\epsilon_i - (x_i - \mu_q)/\sqrt{v_q}\right)v_q/v_g
\end{align}
where we have used $\mathcal{L}_i(\epsilon)$ for short of $\mathcal{L}_i(\epsilon|(x_i - \mu_q)/\sqrt{v_q})$.

It is interesting to note that the approximations of the reparameterized likelihood factors look remarkably similar to the approximate terms $\tilde t_i(\theta)$ in \cite{Minka2001phd}. This is not a coincidence since both have the same functional form of exponentiated quadratic and both approximate likelihood factors. However, $\tilde t_i(\theta)$ is a global approximation and is defined as a scaled ratio between the new variational distribution and the old $\tilde t_i(\theta) = Zq^{new}(\theta)/q(\theta)$ \cite{Minka2001phd}. In contrast, $\mathcal{\widetilde L}_i(\epsilon)$ is a local approximation and is determined by the second order Taylor series expansion of $\ln\mathcal{L}_i(\epsilon)$ at $\epsilon_i$. Furthermore, the two are used in very different contexts and for different purpose.

\subsection{Analytically Solving the Integrals}
\label{subsec:solveintegrals}

Since both $\rho_i(\epsilon)$ and $\mathcal{\widetilde L}_i(\epsilon)$ are exponentiated quadratics, the product $\rho_i(\epsilon)/\mathcal{\widetilde L}_i(\epsilon)$ is also an exponentiated quadratic and after some rearranging can be expressed as follows:
\begin{equation}
q_0(\epsilon)\pi_i(\epsilon) \approx \rho_i(\epsilon)/\mathcal{\widetilde L}_i(\epsilon) = \pi_i(\epsilon_i)\sqrt{\hat v_i}A_i \mathcal{N}(\epsilon;\hat\epsilon_i,\hat v_i)\label{eq:q0piapprox}
\end{equation}
where
\begin{align}
&\hat v_i = \left.v_g\middle/\left[\left(1 - \pi_i(\epsilon_i)\right)\left(\pi_i(\epsilon_i)v_g\left(\frac{x_i - \mu_q}{v_g + v_q}\right)^2 + 1\right)v_q + v_g\right]\right. \\
&\hat\epsilon_i = \sqrt{v_q}\left(1 - \pi_i(\epsilon_i)\hat v_i\right)\frac{x_i - \mu_q}{v_g + v_q}\label{eq:hatepsiloni} \\
&A_i = \exp\left[-\frac{1}{2}\left(1 - \pi_i(\epsilon_i)^2\hat v_i\right)v_q\left(\frac{x_i - \mu_q}{v_g + v_q}\right)^2\right]
\end{align}
It can be verified that $\mathcal{N}(\epsilon;\hat\epsilon_i,\hat v_i)$ is a proper Gaussian because $\hat v_i$ is strictly positive. 

Alternatively, the Laplace approximation can also be used to locally approximate $q_0(\epsilon)\pi_i(\epsilon)$ by finding its mode and computing the variance at the mode. However, finding the mode of $q_0(\epsilon)\pi_i(\epsilon)$ requires in itself iterative approximation, which would add significant complexity to the whole method because it has to be performed for each data point.

Next, to compute the integrals in equation (\ref{eq:elbogradu}) and (\ref{eq:elbogradv}) we substitute $q_0(\epsilon)\pi_i(\epsilon)$ with its approximation from equation (\ref{eq:q0piapprox}). Carrying out the integration over $\epsilon$ and rearranging we obtain:
\begin{align}
&G_{\mu_q} \approx \widetilde G_{\mu_q} = \sum_{i}B_i\frac{x_i - \mu_q}{v_g} + \frac{\mu_p - \mu_q}{v_p} \label{eq:elbogradu2} \\
&G_{v_q} \approx \widetilde G_{v_q} = \frac{1}{2}\left[-\sum_{i}C_i\frac{1}{v_g} + \sum_{i}D_i\frac{(x_i - \mu_q)^2}{v_g}\frac{1}{v_g + v_q} + \frac{1}{v_q} - \frac{1}{v_p}\right]\label{eq:elbogradv2} \\
&B_i = \pi_i(\epsilon_i)\sqrt{\hat v_i}A_i\frac{v_g + \pi_i(\epsilon_i)\hat v_i v_q}{v_g + v_q}\label{eq:defBi} \\
&C_i = \pi_i(\epsilon_i)\sqrt{\hat v_i}A_i\hat v_i\label{eq:defCi} \\
&D_i = (1 - \pi_i(\epsilon_i) \hat v_i)B_i\label{eq:defDi}
\end{align}

Equations (\ref{eq:elbogradu2}) and (\ref{eq:elbogradv2}) represent an analytical approximation of the ELBO gradient over the parameters of the variational distribution and are illustrated in figure \ref{fig:fig2}.

\subsection{Extension to Multidimensional Data}
\label{subsec:extmultidim}

Let $M$ be the dimensionality of the data. Since the data space can be scaled and rotated arbitrarily, we can assume without loss of generality that the multivariate Gaussian in the observation density is spherical. Consequently, the gradient over the mean of the variational distribution is separable and can be broken into $M$ one dimensional gradients along the principal axes of the ellipsoid defining the variational distribution. Similarly, the gradient over the diagonal elements of $\Sigma_q$ is also separable and can be broken into $M$ one dimensional variance gradients along the principal axes. However, efficiently determining the off-diagonal elements of $\Sigma_q$ is not trivial and a possible approach can be to adjust the principal axes by employing principal component analysis in an iterative optimization scheme, essentially breaking the multidimensional problem into multiple single dimensional ones.

\subsection{Applicability to Non-Gaussian Distributions}
\label{subsec:extmultidim}

The main restriction for the applicability of the proposed approximation to non-Gaussian distributions comes from the reparameterization trick, which does not easily generalize to other common distributions, such as the gamma or beta. While a number of methods have been proposed to circumvent that problem (\cite{Ruiz2016}, \cite{Figurnov2018}), these have been developed with stochastic approximation of the expectation in mind and are not readily applicable to analytical integration. A possible solution for this limitation can be to approximate the target distribution with a Gaussian mixture and extend the proposed ELBO gradient approximation to handle Gaussian mixtures.

\section{Maximizing ELBO}
\label{sec:maxelbo}

To test the proposed analytical approximation of the ELBO gradient we integrate it in the expectation step of an EM algorithm for ELBO maximization, as specified in algorithm \ref{alg:elbogaaem}. The approximation allows the expectation step to be applied directly on the ELBO gradient rather than on ELBO itself. In the maximization step we then derive update rules for optimization of the variational distribution parameters $\mu_q$ and $v_q$ through fixed-point iteration. Our aim is to construct linear update rules by keeping constant, with respect to the current values of $\mu_q$ and $v_q$, all terms that are limited in range. For each update rule convergence is briefly examined by demonstrating that the update is in the direction of the local maximum stationary point but we leave a detailed convergence analysis for future work.

Since the approximation relies on the assumption that the variational distribution is more compactly supported than the Gaussian distribution in the likelihood factors, to satisfy that condition during the first few iterations when $v_q$ is large we introduce a substitute for $v_g$, denoted $\hat v_g$, and initialize it to $max(2v_q,v_g)$. The factor $2$ is chosen because it provides acceptable worst case error of the approximation. The substitute $\hat v_g$ is then gradually reduced with each iteration of the algorithm according to the rule $\hat v_g = max(min(v_q,\hat v_g/2),v_g)$ until it reaches the value of $v_g$. The value of $\hat v_g$ is used in place of $v_g$ in all expressions of the gradient approximation and the update rules but for clarity purposes we do not specify it explicitly. In addition, to prevent $v_q$ from breaking the assumption of compact support it is constrained by the rule $v_q = min(v_q,max(v_g,\hat v_g/2))$ at the end of each iteration. 

\begin{algorithm}[t!]
\caption{EM with analytical approximation of the ELBO gradient}\label{alg:elbogaaem}
{
\begin{enumerate}
  \item Input: data $x_i$, size $n$, $w$, $Pc(x_i)$, $v_g$, $\mu_p$, $v_p$
  \item Initialize $\mu_q = \sum_{i}x_i/n$, $v_q = \sum_{i}(x_i - \mu_q)^2/n + v_g$, $\hat v_g = max(2v_q,v_g)$
  \item For $t = 1$ to convergence or maximum number of iterations
    \begin{enumerate}
       \item Expectation step: approximate the ELBO gradient and compute $\pi_i(\epsilon_i)$, $\hat v_i$,  $A_i$, $B_i$, $C_i$ and $D_i$
       \item Maximization step: update the parameters of the variational distribution $\mu_q$ and $v_q$ according to the update rules
       \item update $\hat v_g = max(min(2v_q,\hat v_g/2),v_g)$, constrain $v_q = min(v_q,max(v_g,\hat v_g/2))$
    \end{enumerate}
\end{enumerate}
}
\end{algorithm}

\subsection{Update Rule for the Mean}
\label{updatemean}

\begin{figure}[t!]
\includegraphics[width=1.0\linewidth]{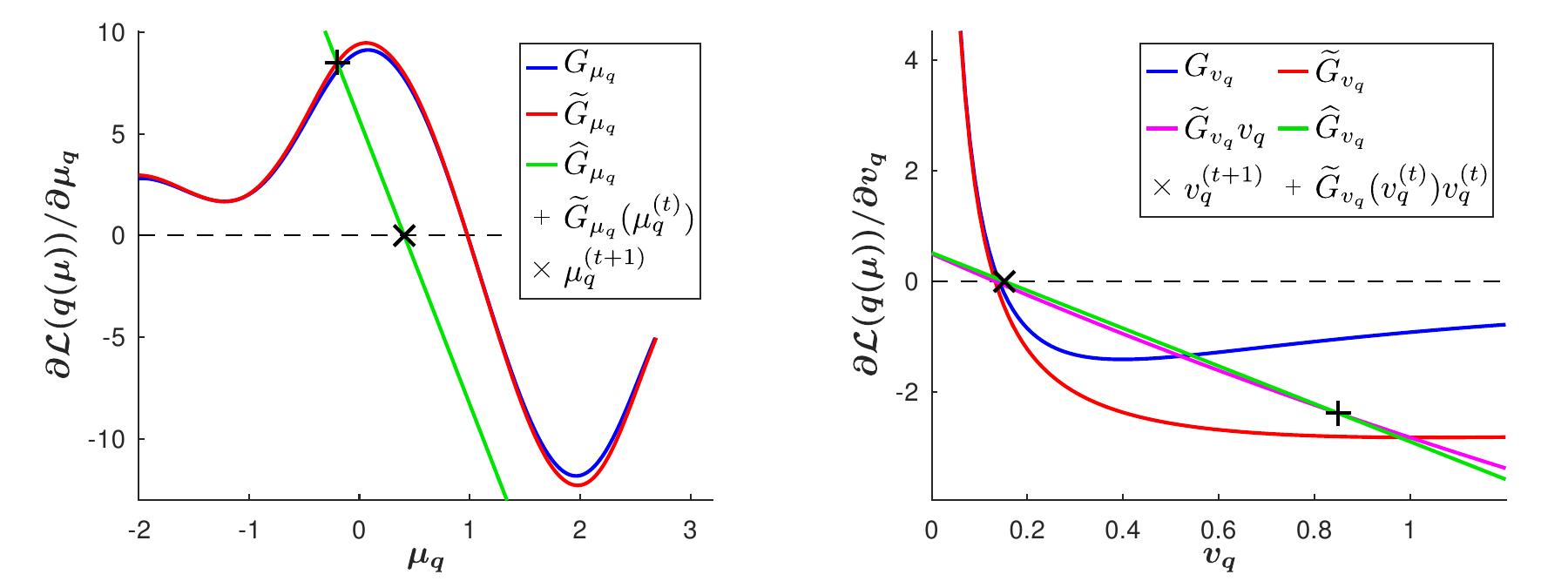}
\caption{Analytical approximation of the ELBO gradient over the mean $\mu_q$ (left panel) and variance $v_q$ (right panel) of the variational distribution}
\label{fig:fig2}
\end{figure}

We start by examining the terms that make up $B_i$ as defined in equation (\ref{eq:defBi}). These are $\pi_i(\epsilon_i)$, $\sqrt{\hat v_i}$, $A_i$ and $(v_g + \pi_i(\epsilon_i)\hat v_i v_q)/(v_g + v_q)$. If we assume that the probability for a non-clutter data point is greater than zero, it can be readily deduced from its definition in equation (\ref{eq:pi_i}) that the term $\pi_i(\epsilon_i)$ is limited to the range $(0,1]$. It follows then that $\hat v_i$ falls within the range $(0,1]$ and therefore so does $\sqrt{\hat v_i}$. Since $\pi_i(\epsilon_i) \in (0,1]$ and $\hat v_i \in (0,1]$, it follows that $(1 - \pi_i(\epsilon_i)^2\hat v_i) \in [0,1)$ and therefore $A_i \in (0,1]$ because it is an exponent with a negative argument. Since $\pi_i(\epsilon_i) \in (0,1]$ and $\hat v_i \in (0,1]$, it also follows that $(v_g + \pi_i(\epsilon_i)\hat v_i v_q)/(v_g + v_q) \in (0,1]$. Consequently, since $B_i$ is a product of the above terms, it follows that $B_i$ is limited to the range $(0,1]$. Therefore, to derive the update rule for the mean $\mu_q$ we keep $B_i$ constant with respect to the current value of $\mu_q$. Setting the approximate gradient $\widetilde G_{\mu_q}$ equal to zero in equation (\ref{eq:elbogradu2}) we then solve the resulting equation for $\mu_q$:
\begin{equation} \label{eq:uqupdate}
\mu_q^{(t+1)} = \frac{\sum_{i}B_i(\mu_q^{(t)})x_i/v_g + \mu_p/v_p}{\sum_{i}B_i(\mu_q^{(t)})/v_g + 1/v_p}
\end{equation}
where $\mu_q^{(t+1)}$ is the updated value and $\mu_q^{(t)}$ is the current value. The corresponding update function is presented in figure \ref{fig:fig2} and is defined by the linear expression:
\begin{equation} \label{eq:hatGuq}
\widehat G_{\mu_q} = \sum_{i}B_i(\mu_q^{(t)})\frac{x_i - \mu_q}{v_g} + \frac{\mu_p - \mu_q}{v_p}
\end{equation}
To confirm that $\mu_q$ is updated in the direction of the local maximum we demonstrate that the slope of the update function $\widehat G_{\mu_q}$ is negative. Let $\mu_q^*$ be the local maximum stationary point ($\widetilde G_{\mu_q}(\mu_q^*) = 0$ and $\partial\widetilde G_{\mu_q}(\mu_q)/\partial\mu_q|_{\mu_q = \mu_q^*} < 0$). If $\widetilde G_{\mu_q}(\mu_q^{(t)}) > 0$ it follows that $\mu_q^* > \mu_q^{(t)}$, since $\widetilde G_{\mu_q}$ is the gradient. At the same time, if $\widetilde G_{\mu_q}(\mu_q^{(t)}) > 0$ and the slope of $\widehat G_{\mu_q}$ is negative it follows that $\mu_q^{(t+1)} > \mu_q^{(t)}$, and therefore the update is in the direction of the local maximum. Similarly, if $\widetilde G_{\mu_q}(\mu_q^{(t)}) < 0$ it follows that $\mu_q^* < \mu_q^{(t)}$ and $\mu_q^{(t+1)} < \mu_q^{(t)}$, and therefore the update is again in the direction of the local maximum. Since we already established that $B_i$ is positive, it follows from equation (\ref{eq:hatGuq}) that the slope of $\widehat G_{\mu_q}$ is negative and therefore $\mu_q$ is always updated in the direction of the local maximum.

\subsection{Update Rule for the Variance}
\label{updatevariance}

We start by multiplying equation (\ref{eq:elbogradv2}) with $v_q$ to eliminate $v_q$ in the denominator. The resulting function $\widetilde G_{v_q}v_q$ is illustrated in figure \ref{fig:fig2} and can be expressed as follows:
\begin{equation}
\widetilde G_{v_q}v_q = \frac{1}{2}\left[-\left(\sum_{i}C_i\frac{1}{v_g} + \frac{1}{v_p}\right)v_q + \sum_{i}D_i\frac{(x_i - \mu_q)^2}{v_g}\frac{v_q}{v_g + v_q} + 1\right]
\end{equation}
Next, we examine the terms $C_i$ and $D_i$. Since we already established that $\pi_i(\epsilon_i) \in (0,1]$, $\hat v_i \in (0,1]$ and $A_i \in (0,1]$, it follows from the definition of $C_i$ in equation (\ref{eq:defCi}) as a product of these terms that $C_i \in (0,1]$. Since $\pi_i(\epsilon_i) \in (0,1]$, $\hat v_i \in (0,1]$ and $B_i \in (0,1]$, it follows that $(1 - \pi_i(\epsilon_i)\hat v_i) \in [0,1)$ and consequently $D_i \in [0,1)$. In addition, the term $v_q/(v_g + v_q)$ is also limited to the range $[0,1)$. Therefore, to derive the update rule for the variance $v_q$ we keep the terms $B_i$, $C_i$, $D_i$ and $v_q/(v_g + v_q)$ constant with respect to the current value of $v_q$. Setting the approximate gradient $\widetilde G_{v_q}$ equal to zero, we then solve the resulting linear equation, obtaining the following update rule for the variance:
\begin{equation} \label{eq:vqupdate}
v_q^{(t+1)} = \left.\left(\sum_{i}D_i(v_q^{(t)})\frac{(x_i - \mu_q)^2}{v_g}\frac{v_q^{(t)}}{v_g + v_q^{(t)}} + 1\right)\middle/\left(\sum_{i}C_i(v_q^{(t)})\frac{1}{v_g} + \frac{1}{v_p}\right)\right.
\end{equation}
where $v_q^{(t+1)}$ is the updated value and $v_q^{(t)}$ is the current value. The corresponding update function is presented in figure \ref{fig:fig2} and is formulated by the linear expression:
\begin{equation} \label{eq:hatGvq}
\widehat G_{v_q} = \frac{1}{2}\left[-\left(\sum_{i}C_i(v_q^{(t)})\frac{1}{v_g} + \frac{1}{v_p}\right)v_q + \sum_{i}D_i(v_q^{(t)})\frac{(x_i - \mu_q)^2}{v_g}\frac{v_q^{(t)}}{v_g + v_q^{(t)}} + 1\right]
\end{equation}

To confirm that $v_q$ is updated in the direction of the local maximum we verify that the slope of $\widehat G_{v_q}$ is negative. Since we already established that $C_i$ is positive, it follows from equation (\ref{eq:hatGvq}) that the slope of $\widehat G_{v_q}$ is negative and therefore $v_q$ is always updated in the direction of the local maximum.

In addition, we demonstrate that equation (\ref{eq:vqupdate}) produces valid updates ($v_q^{(t+1)} > 0$) by examining the update function at $v_q = 0$. In that case the term $(\sum_{i}C_i(v_q^{(t)})/v_g + 1/v_p)v_q$ is equal to zero because $C_i(v_q^{(t)})$ is finite. Since $D_i \in [0,1)$ and $v_q/(v_g + v_q) \in [0,1)$, it follows that the term $(\sum_{i}D_i(v_q^{(t)})(x_i - \mu_q)^2/v_g)v_q^{(t)}/(v_g + v_q^{(t)}) \geqslant 0$. Consequently, the update function $\widehat G_{v_q}$ is strictly positive at $v_q = 0$ and since, as already established, its slope is negative, it follows that the root $v_q^{(t+1)}$ is strictly positive and therefore is a valid update for $v_q$.

\section{Comparative Testing}
\label{sec:comparetest}

\begin{figure}[t!]
\includegraphics[width=1.0\linewidth]{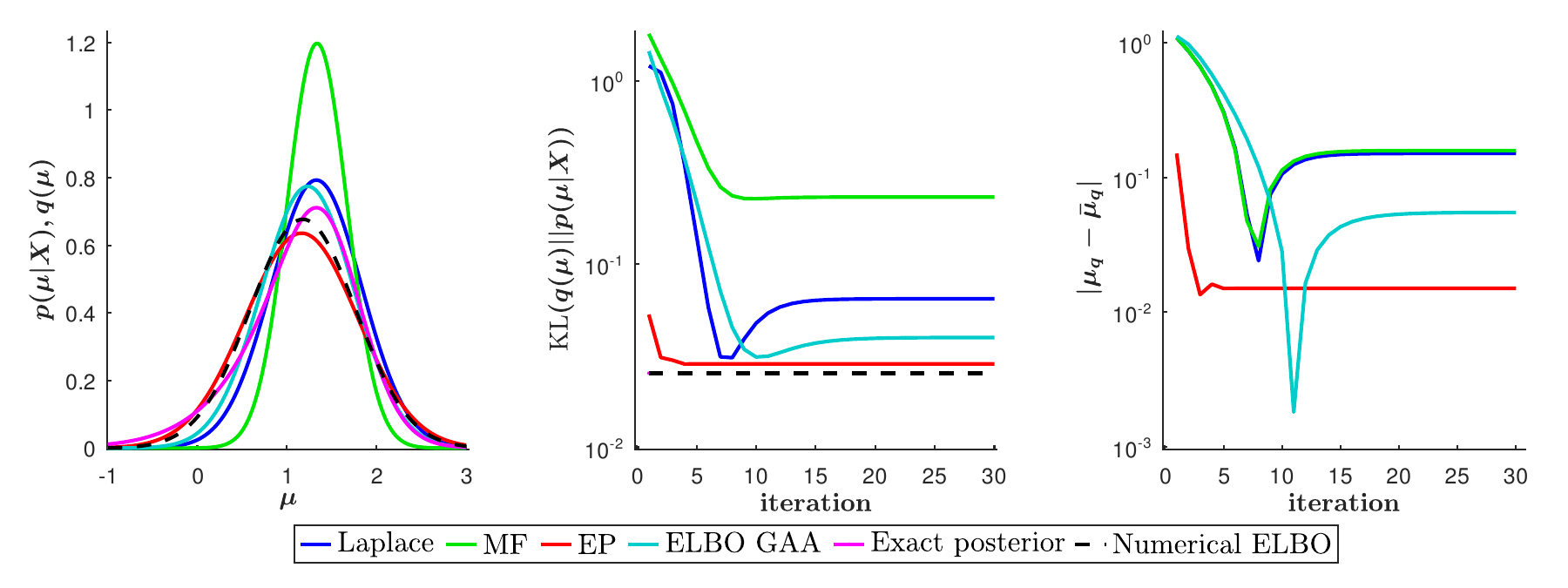}
\caption{Testing the ELBO gradient analytical approximation (ELBO GAA) against classical deterministic approaches for a sample size of 20 data points}
\label{fig:fig3}
\end{figure}

We test the proposed method for analytical approximation of the ELBO gradient against classical deterministic approaches such as the Laplace approximation, expectation propagation and mean-field variational inference. Our implementation of the Laplace approximation and mean-field variational inference is based on the description of the algorithms in \cite{Bishop2006} and for expectation propagation we use \cite{Minka2001}. The test data is generated from the clutter problem observation density as defined in equation (\ref{eq:observation}), for one dimensional data and using the same parameters as specified in \cite{Minka2001phd}: $w = 0.5$, $Pc(x) = \mathcal{N}(x;\mu_c = 0,v_c = 10)$, Gaussian distribution in the observation density $\mathcal{N}(x;\mu = 2,v_g = 1)$, number of data points $n = 20$ and we also assume the same prior distribution $p(\mu) = \mathcal{N}(\mu;\mu_p = 0,v_p = 100)$. The aim is to determine how well the proposed method approximates the posterior distribution $p(\mu|X)$ in comparison to the other three methods. As a measure of goodness of approximation we use the KL divergence between the variational distribution and the posterior, $\text{KL}(q(\mu)||p(\mu|X)) = \ln p(X) - \mathcal{L}(q(\mu))$. This is calculated by numerically evaluating the log marginal likelihood $\ln p(X)$ and the evidence lower bound $\mathcal{L}(q(\mu))$ for each method. In addition, a baseline of best achievable approximation is provided for reference by numerically maximizing ELBO. We denote the mean of the baseline variational distribution $\bar \mu_q$ and use it to define absolute error rates $|\mu_q - \bar \mu_q|$ for the means of the tested variational distributions.

\begin{figure}[t!]
\includegraphics[width=1.0\linewidth]{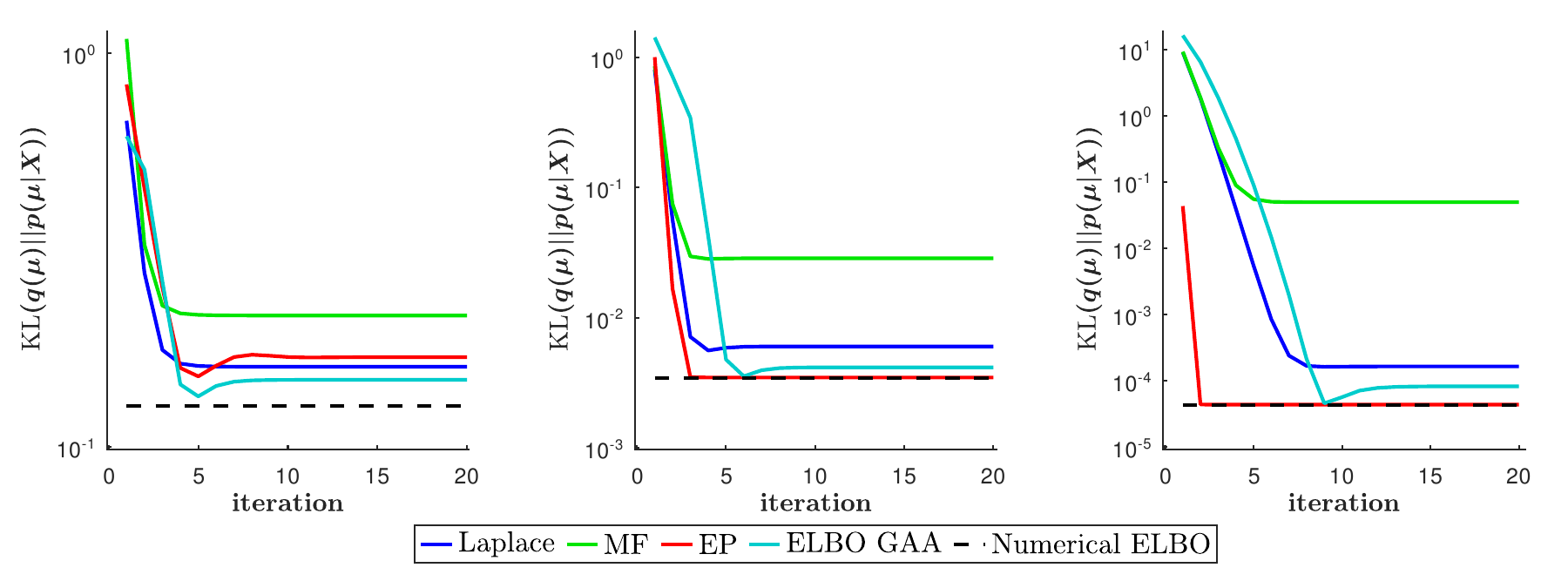}
\caption{Testing the proposed approximation (ELBO GAA) against classical deterministic approaches for sample sizes of 5 (left), 10 (middle) and 100 (right) data points}
\label{fig:fig4}
\end{figure}

The result of one test with 20 data points is presented in figure \ref{fig:fig3}. A data sample with a rather skewed posterior is selected to give the tested methods a challenge. To summarize the test results, the proposed method scores second behind EP in terms of KL divergence and absolute error of the mean, beating both mean-field variational inference and the Laplace approximation by a significant margin. It also demonstrates a good rate of convergence, slightly worse than MF and Laplace. Furthermore, in contrast to EP, which can fail to converge (\cite{Minka2001}) or in certain circumstances can produce negative $v_q$ (\cite{Minka2001phd}), the proposed method appears to not suffer from problems with convergence and as demonstrated in section \ref{updatevariance} is guaranteed to have strictly positive $v_q$.

An interesting observation is that just before convergence is reached the KL divergence of the proposed method increases slightly. This is easily explained. The optimization of the variational distribution parameters is performed with respect to the local maximum stationary point defined by the approximate ELBO gradient, which is slightly different from the stationary point of the real ELBO gradient due to the imperfection of the approximation. As a result, when the parameters are far away from the two stationary points, updating them towards the appproximate stationary point means also a decrease in the distance to the real stationary point and hence a reduction of the KL divergence. However, when the parameters are very close to the real stationary point updating them towards the appproximate stationary point may in some cases increase the distance to the real stationary point and hence increase the KL divergence.

Figure \ref{fig:fig4} illustrates the performance of the tested algorithms for 5, 10 and 100 data points. As can be observed, the proposed method performs well for all three cases and for the case of 5 data points even comes out the best in terms of KL divergence.

\begin{figure}[t!]
\includegraphics[width=1.0\linewidth]{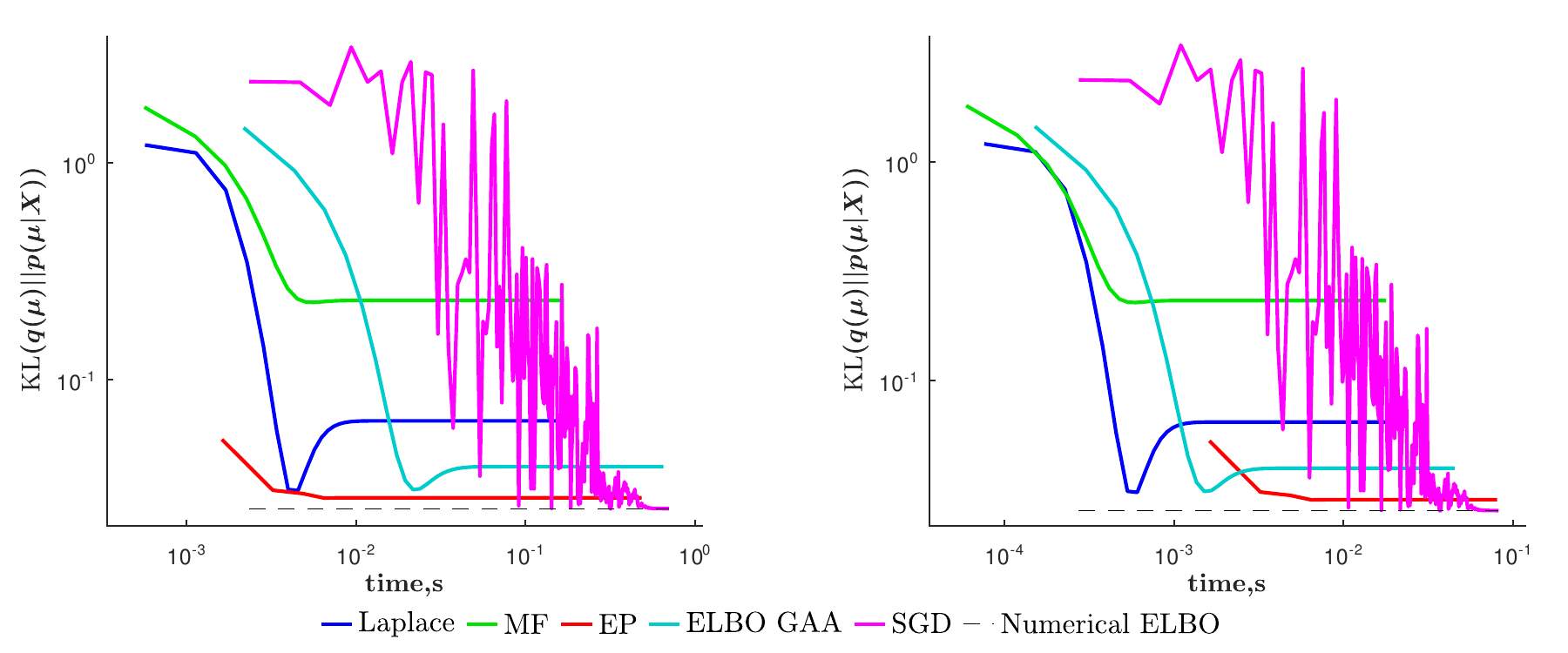}
\caption{Execution times for the proposed method (ELBO GAA) in comparison to classical deterministic approaches and stochastic gradient descent (SGD) for 20 data points with serial execution (left panel) and vectorization (right panel)}
\label{fig:fig5}
\end{figure}

In terms of computational complexity, the proposed method is linearly dependent on the number of data points $n$ just like the other three methods. However, the number of exponentiation operations is $2n$ versus $n$ for EP, MF and Laplace. Furthermore, there are also $n$ square roots and the number of arithmetic operations is generally larger. An advantage of the proposed method is that it is a parallel algorithm and can be easily vectorized, provided the hardware has a vector processing unit. This is in contrast to EP, which is a serial algorithm, while MF and Laplace are also parallel algorithms. To give a rough illustration of computational requirements we measure execution times for 20 data points on a standard desktop machine within the Octave environment with serial execution and vectorization. The results are presented in figure \ref{fig:fig5} and also include stochastic gradient descent (SGD) as a point of comparison to stochastic methods. SGD is implemented using the ELBO gradient by employing the reparameterization trick and is vectorizable as well. The hyperparameters of SGD are tuned for achieving fast convergence where we use 3 samples to compute the gradient expectation and the learning rate starts at 1.0 with a decay factor of 0.97.

\section{Conclusion}
\label{sec:conclusion}

We describe a method for analytical approximation of the ELBO gradient in the context of the clutter problem. The approximation is integrated into the expectation step of an EM algorithm for ELBO maximization and update rules for optimizing the variational parameters are derived. We test the proposed method against classical deterministic approaches such as the Laplace approximation, expectation propagation and mean-field variational inference, and the method demonstrates good accuracy and rate of convergence for both small and large number of data points. The computational complexity is linearly dependent on the number of data points, but the number of mathematical operations is somewhat larger compared to the other three algorithms in the test. We provide limited convergence analysis for the update rules of the developed EM algorithm and, while empirical data suggests convergence is reliable, a detailed analytical proof of convergence is a primary research task for the future. Another area where the proposed method requires further work is the extension to multidimensional data observations and the handling of non-Gaussian distributions. While the proposed method provides a solution for the maximization of ELBO by optimizing the variational distribution parameters it does not offer a means to calculate the actual value of ELBO and as a result can not be used for model selection.


\appendix

\section{Approximating the Terms $\boldsymbol{q_0(\epsilon)\pi_i(\epsilon)}$ with a Gaussian}\label{apd:first}

Substituting $\pi_i(\epsilon)$ with its definition from equation (\ref{eq:pi_i}) we have:
\begin{equation}
q_0(\epsilon)\pi_i(\epsilon) = q_0(\epsilon)\frac{(1 - w)\mathcal{N}(\epsilon;(x_i - \mu_q)/\sqrt{v_q},v_g/v_q)/\sqrt{v_q}}{(1 - w)\mathcal{N}(\epsilon;(x_i - \mu_q)/\sqrt{v_q},v_g/v_q)/\sqrt{v_q} + wP_c(x_i)}
\end{equation}
Since both $q_0(\epsilon)$ and $(1 - w)\mathcal{N}(\epsilon;(x_i - \mu_q)/\sqrt{v_q},v_g/v_q)/\sqrt{v_q}$ are Gaussian, their product is conveniently also Gaussian, wich we denote $\rho_i(\epsilon)$:
\begin{equation} \label{eq:rho0}
\rho_i(\epsilon) = (1 - w)\frac{1}{\sqrt{2(1 + v_g/v_q)v_q\pi}}\exp\left(-\frac{1}{2}\frac{(0 - \left(x_i - \mu_q)/\sqrt{v_q}\right)^2}{1 + v_g/v_q}\right)\mathcal{N}(\epsilon;\epsilon_i,v_i)
\end{equation}
where $\epsilon_i = \sqrt{v_q}(x_i - \mu_q)/(v_g + v_q)$ and $v_i = v_g/(v_g + v_q)$. Rearranging and simplifying equation (\ref{eq:rho0}) we obtain:
\begin{equation}
\rho_i(\epsilon) = (1 - w)\frac{1}{\sqrt{2(v_g + v_q)\pi}}\exp\left(-\frac{1}{2}\frac{(x_i - \mu_q)^2}{v_g + v_q}\right)\mathcal{N}(\epsilon;\epsilon_i,v_i)
\end{equation}

The denominator of $\pi_i(\epsilon)$ is the reparameterized likelihood factor $\mathcal{L}_i(\epsilon) = (1 - w)\mathcal{N}(\epsilon;(x_i - \mu_q)/\sqrt{v_q},v_g/v_q)/\sqrt{v_q} + wP_c(x_i)$ and, because of the assumption for compact support of the variational distribution, it can be efficiently approximated locally at the point $\epsilon_i$ with an exponentiated quadratic, where the quadratic is derived by a second order Taylor series expansion of $\ln\mathcal{L}_i(\epsilon)$ at $\epsilon_i$:
\begin{align}
&\mathcal{L}_i(\epsilon) \approx \mathcal{\widetilde L}_i(\epsilon) = \exp\left(\frac{1}{2}\ln(\mathcal{L}_i(\epsilon))''|_{\epsilon = \epsilon_i}(\epsilon - \epsilon_i)^2 + \ln(\mathcal{L}_i(\epsilon))'|_{\epsilon = \epsilon_i}(\epsilon - \epsilon_i) + \ln\mathcal{L}_i(\epsilon_i)\right)\label{eq:tildeLi} \\
&\ln(\mathcal{L}_i(\epsilon))''|_{\epsilon = \epsilon_i} = \pi_i(\epsilon_i)\left(\left(1 - \pi_i(\epsilon_i)\right)\left(\epsilon_i - (x_i - \mu_q)/\sqrt{v_q}\right)^2v_q/v_g - 1\right)v_q/v_g \\
&\ln(\mathcal{L}_i(\epsilon))'|_{\epsilon = \epsilon_i} = -\pi_i(\epsilon_i)\left(\epsilon_i - (x_i - \mu_q)/\sqrt{v_q}\right)v_q/v_g
\end{align}
Deriving $\ln(\mathcal{L}_i(\epsilon))'$ is straightforward and for $\ln(\mathcal{L}_i(\epsilon))''$ we use the product rule of derivatives:
\begin{equation} \label{eq:d2Li}
(\ln(\mathcal{L}_i(\epsilon))')' = -\left(\pi_i(\epsilon)'\left(\epsilon - (x_i - \mu_q)/\sqrt{v_q}\right) + \pi_i(\epsilon)\left(\epsilon - (x_i - \mu_q)/\sqrt{v_q}\right)'\right)v_q/v_g
\end{equation}
We then apply the product rule of derivatives also to $\pi_i(\epsilon)'$ and rearrange:
\begin{equation}
\pi_i(\epsilon)' = -\pi_i(\epsilon_i)(1 - \pi_i(\epsilon_i))\left(\epsilon - (x_i - \mu_q)/\sqrt{v_q}\right)v_q/v_g
\end{equation}
Replacing $\pi_i(\epsilon)'$ back in equation (\ref{eq:d2Li}) we obtain:
\begin{align}
&\ln(\mathcal{L}_i(\epsilon))'' = -\left(-\pi_i(\epsilon_i)(1 - \pi_i(\epsilon_i))\left(\epsilon - (x_i - \mu_q)/\sqrt{v_q}\right)^2v_q/v_g + \pi_i(\epsilon)\right)v_q/v_g \\
&\ln(\mathcal{L}_i(\epsilon))'' = \pi_i(\epsilon)\left(\left(1 - \pi_i(\epsilon)\right)\left(\epsilon - (x_i - \mu_q)/\sqrt{v_q}\right)^2v_q/v_g - 1\right)v_q/v_g
\end{align}
Next we substitute $\epsilon_i$ in the numerator of $\pi_i(\epsilon_i)$ with its definition $\sqrt{v_q}(x_i - \mu_q)/(v_g + v_q)$ and simplify the resulting expression:
\begin{align}
&\frac{(1 - w)}{\sqrt{v_q}}\mathcal{N}\left(\epsilon_i;\frac{(x_i - \mu_q)}{\sqrt{v_q}},\frac{v_g}{v_q}\right) = \frac{(1 - w)}{\sqrt{2v_g\pi}}\exp\left(-\frac{1}{2}\left(\sqrt{v_q}\frac{x_i - \mu_q}{v_g + v_q} - \frac{x_i - \mu_q}{\sqrt{v_q}}\right)^2\frac{v_q}{v_g}\right) \\
&\frac{(1 - w)}{\sqrt{v_q}}\mathcal{N}\left(\epsilon_i;\frac{(x_i - \mu_q)}{\sqrt{v_q}},\frac{v_g}{v_q}\right) = \frac{(1 - w)}{\sqrt{2v_g\pi}}\exp\left(-\frac{1}{2}\left(\frac{v_q}{v_g + v_q} - 1\right)^2\frac{(x_i - \mu_q)^2}{v_g}\right) \\
&\frac{(1 - w)}{\sqrt{v_q}}\mathcal{N}\left(\epsilon_i;\frac{(x_i - \mu_q)}{\sqrt{v_q}},\frac{v_g}{v_q}\right) = (1 - w)\frac{1}{\sqrt{2v_g\pi}}\exp\left(-\frac{1}{2}\frac{v_g(x_i - \mu_q)^2}{(v_g + v_q)^2}\right)\label{eq:numeratorpi}
\end{align}
We then rewrite equation (\ref{eq:tildeLi}) by completing the square inside the exponent, where for simplicity we have defined $a_i = \ln(\mathcal{L}_i(\epsilon))''|_{\epsilon = \epsilon_i}$ and $b_i = \ln(\mathcal{L}_i(\epsilon))'|_{\epsilon = \epsilon_i}$:
\begin{equation}
\mathcal{\widetilde L}_i(\epsilon) = \exp\left(\frac{1}{2}\frac{\left(\epsilon - (\epsilon_i - b_i/a_i)\right)^2}{1/a_i}\right)\exp\left(-\frac{1}{2}\frac{b_i^2}{a_i}\right)\mathcal{L}_i(\epsilon_i)
\end{equation}
The product $\rho_i(\epsilon)/\mathcal{\widetilde L}_i(\epsilon)$ can be formulated then as follows:
\begin{align}
\frac{\rho_i(\epsilon)}{\mathcal{\widetilde L}_i(\epsilon)} = &(1 - w)\frac{1}{\sqrt{2(v_g + v_q)\pi}}\exp\left(-\frac{1}{2}\frac{(x_i - \mu_q)^2}{v_g + v_q}\right)\frac{1}{\sqrt{2v_i\pi}}\exp\left(-\frac{1}{2}\frac{(\epsilon - \epsilon_i)^2}{v_i}\right) \nonumber \\
&\exp\left(-\frac{1}{2}\frac{\left(\epsilon - (\epsilon_i - b_i/a_i)\right)^2}{1/a_i}\right)\exp\left(\frac{1}{2}\frac{b_i^2}{a_i}\right)/\mathcal{L}_i(\epsilon_i)
\end{align}
Substituting $v_i$ in $\sqrt{2v_i\pi}$ with $v_g/(v_g + v_q)$ we cancell out the terms $(v_g + v_q)$ and substituting $\exp\left(-(1/2)(x_i - \mu_q)^2/(v_g + v_q)\right)$ with $\exp\left(-(1/2)(x_i - \mu_q)^2(v_g + v_q)/(v_g + v_q)^2\right)$ the expression for the product $\rho_i(\epsilon)/\mathcal{\widetilde L}_i(\epsilon)$ becomes:
\begin{align}
\frac{\rho_i(\epsilon)}{\mathcal{\widetilde L}_i(\epsilon)} = &(1 - w)\frac{1}{\sqrt{2\pi}}\frac{1}{\sqrt{2v_g\pi}}\exp\left(-\frac{1}{2}\frac{v_g(x_i - \mu_q)^2}{(v_g + v_q)^2}\right)\exp\left(-\frac{1}{2}\frac{v_q(x_i - \mu_q)^2}{(v_g + v_q)^2}\right)\nonumber \\
&\exp\left(-\frac{1}{2}\frac{(\epsilon - \epsilon_i)^2}{v_i}\right)\exp\left(-\frac{1}{2}\frac{\left(\epsilon - (\epsilon_i - b_i/a_i)\right)^2}{1/a_i}\right)\exp\left(\frac{1}{2}\frac{b_i^2}{a_i}\right)/\mathcal{L}_i(\epsilon_i)
\end{align}
Since $(1 - w)\left(1/\sqrt{2v_g\pi}\right)\exp\left(-(1/2)v_g(x_i - \mu_q)^2/(v_g + v_q)^2\right)$ is the numerator of $\pi_i(\epsilon_i)$ as defined in equation (\ref{eq:numeratorpi}) and $\mathcal{L}_i(\epsilon_i)$ is the denominator, the product $\rho_i(\epsilon)/\mathcal{\widetilde L}_i(\epsilon)$ is simplified to the following:
\begin{align}
\frac{\rho_i(\epsilon)}{\mathcal{\widetilde L}_i(\epsilon)} = &\frac{1}{\sqrt{2\pi}}\pi_i(\epsilon_i)\exp\left(-\frac{1}{2}\frac{v_q(x_i - \mu_q)^2}{(v_g + v_q)^2} + \frac{1}{2}\frac{b_i^2}{a_i}\right)\nonumber \\
&\exp\left(-\frac{1}{2}\frac{(\epsilon - \epsilon_i)^2}{v_i}\right)\exp\left(-\frac{1}{2}\frac{\left(\epsilon - (\epsilon_i - b_i/a_i)\right)^2}{1/a_i}\right)
\end{align}
We then carry out the multiplication of the two exponentiated quadratics:
\begin{align}
&\frac{\rho_i(\epsilon)}{\mathcal{\widetilde L}_i(\epsilon)} = \frac{1}{\sqrt{2\pi}}\pi_i(\epsilon_i)A_i\exp\left(-\frac{1}{2}\frac{(\epsilon - \hat\epsilon_i)^2}{\hat v_i}\right)\label{eq:rhopi0} \\
&\hat v_i = v_i(1/a_i)/(v_i + 1/a_i)\label{eq:hatvi0} \\
&\hat\epsilon_i = (\epsilon_i/a_i + (\epsilon_i - b_i/a_i)v_i)/(v_i + 1/a_i)\label{eq:hatei0} \\
&A_i = \exp\left[-\frac{1}{2}\left(\frac{v_q(x_i - \mu_q)^2}{(v_g + v_q)^2} - \frac{b_i^2}{a_i} + \frac{b_i^2}{a_i^2(v_i + 1/a_i)}\right)\right]\label{eq:Ai0}
\end{align}
Expanding and rearranging equations (\ref{eq:hatvi0}), (\ref{eq:hatei0}) and (\ref{eq:Ai0}) we obtain:
\begin{align}
&\hat v_i = 1/(a_i + 1/v_i) \\
&\hat v_i = v_g/(a_iv_g + v_q + v_g) \\
&\hat v_i = \left.v_g\middle/\left[\pi_i(\epsilon_i)\left(\left(1 - \pi_i(\epsilon_i)\right)\left(\epsilon_i - (x_i - \mu_q)/\sqrt{v_q}\right)^2v_q/v_g - 1\right)v_q + v_q + v_g\right]\right. \\
&\hat v_i = \left.v_g\middle/\left[\left(1 - \pi_i(\epsilon_i)\right)\left(\pi_i(\epsilon_i)\left(\epsilon_i - (x_i - \mu_q)/\sqrt{v_q}\right)^2v_q/v_g + 1\right)v_q + v_g\right]\right. \\
&\hat v_i = \left.v_g\middle/\left[\left(1 - \pi_i(\epsilon_i)\right)\left(\pi_i(\epsilon_i)(x_i - \mu_q)^2\left(v_q/(v_g + v_q) - 1\right)^2/v_g + 1\right)v_q + v_g\right]\right. \\
&\hat v_i = \left.v_g\middle/\left[\left(1 - \pi_i(\epsilon_i)\right)\left(\pi_i(\epsilon_i)v_g(x_i - \mu_q)^2/(v_g + v_q)^2 + 1\right)v_q + v_g\right]\right.
\end{align}
\begin{align}
&\hat\epsilon_i = (\epsilon_i/v_i + \epsilon_ia_i - b_i)/(a_i + 1/v_i) \\
&\hat\epsilon_i = \epsilon_i - b_i\hat v_i \\
&\hat\epsilon_i = \epsilon_i + \pi_i(\epsilon_i)\hat v_i\left(\epsilon_i - (x_i - \mu_q)/\sqrt{v_q}\right)v_q/v_g \\
&\hat\epsilon_i = \left(\sqrt{v_q} + \pi_i(\epsilon_i)\hat v_i\left(\sqrt{v_q} - (v_g + v_q)/\sqrt{v_q}\right)v_q/v_g\right)(x_i - \mu_q)/(v_g + v_q) \\
&\hat\epsilon_i = \sqrt{v_q}\left(1 - \pi_i(\epsilon_i)\hat v_i\right)(x_i - \mu_q)/(v_g + v_q)
\end{align}
\begin{align}
&A_i = \exp\left[-\frac{1}{2}\left(\frac{v_q(x_i - \mu_q)^2}{(v_g + v_q)^2} + \frac{b_i^2/a_i^2 - b_i^2(v_i + 1/a_i)/a_i}{v_i + 1/a_i}\right)\right] \\
&A_i = \exp\left[-\frac{1}{2}\left(\frac{v_q(x_i - \mu_q)^2}{(v_g + v_q)^2} - \frac{b_i^2v_i/a_i}{v_i + 1/a_i}\right)\right] \\
&A_i = \exp\left[-\frac{1}{2}\left(\frac{v_q(x_i - \mu_q)^2}{(v_g + v_q)^2} - b_i^2\hat v_i\right)\right] \\
&A_i = \exp\left[-\frac{1}{2}\left(\frac{v_q(x_i - \mu_q)^2}{(v_g + v_q)^2} - \pi_i(\epsilon_i)^2\left(\frac{\sqrt{v_q}(x_i - \mu_q)}{v_g + v_q} - \frac{x_i - \mu_q}{\sqrt{v_q}}\right)^2v_q^2/v_g^2\hat v_i\right)\right] \\
&A_i = \exp\left[-\frac{1}{2}\left(\frac{v_q(x_i - \mu_q)^2}{(v_g + v_q)^2} - \pi_i(\epsilon_i)^2\left(\frac{v_q}{v_g + v_q} - 1\right)^2v_q/v_g^2\hat v_i(x_i - \mu_q)^2\right)\right] \\
&A_i = \exp\left[-\frac{1}{2}\left(\frac{v_q(x_i - \mu_q)^2}{(v_g + v_q)^2} - \pi_i(\epsilon_i)^2\hat v_i\frac{v_q(x_i - \mu_q)^2}{(v_g + v_q)^2}\right)\right] \\
&A_i = \exp\left[-\frac{1}{2}\left(1 - \pi_i(\epsilon_i)^2\hat v_i\right)v_q\left(\frac{x_i - \mu_q}{v_g + v_q}\right)^2\right]
\end{align}
Since $0 \leqslant \pi_i(\epsilon_i) \leqslant 1$, it follows that $0 < \hat v_i \leqslant 1$ and therefore equation (\ref{eq:rhopi0}) represents a proper Gaussian which results in the final expression for $\rho_i(\epsilon)/\mathcal{\widetilde L}_i(\epsilon)$ as a Gaussian approximation of $q_0(\epsilon)\pi_i(\epsilon)$:
\begin{align}
&q_0(\epsilon)\pi_i(\epsilon) \approx \rho_i(\epsilon)/\mathcal{\widetilde L}_i(\epsilon) = \pi_i(\epsilon_i)\sqrt{\hat v_i}A_i \mathcal{N}(\epsilon;\hat\epsilon_i,\hat v_i)\label{eq:q0piapprox1} \\
&\hat v_i = \left.v_g\middle/\left[\left(1 - \pi_i(\epsilon_i)\right)\left(\pi_i(\epsilon_i)v_g\left(\frac{x_i - \mu_q}{v_g + v_q}\right)^2 + 1\right)v_q + v_g\right]\right. \\
&\hat\epsilon_i = \sqrt{v_q}\left(1 - \pi_i(\epsilon_i)\hat v_i\right)\frac{x_i - \mu_q}{v_g + v_q} \\
&A_i = \exp\left[-\frac{1}{2}\left(1 - \pi_i(\epsilon_i)^2\hat v_i\right)v_q\left(\frac{x_i - \mu_q}{v_g + v_q}\right)^2\right]
\end{align}

\section{Solving the Integrals}\label{apd:second}

To compute the integral in equation (\ref{eq:elbogradu}) we substitute $q_0(\epsilon)\pi_i(\epsilon)$ with its approximation from equation (\ref{eq:q0piapprox}):
\begin{equation}
G_{\mu_q} \approx \widetilde G_{\mu_q} = -\sqrt{v_q}\sum_{i}\pi_i(\epsilon_i)\sqrt{\hat v_i}A_i\int\mathcal{N}(\epsilon;\hat\epsilon_i,\hat v_i)\frac{\epsilon - (x_i - \mu_q)/\sqrt{v_q}}{v_g}d\epsilon + \frac{(\mu_p - \mu_q)}{v_p}
\end{equation}
Modifying the term $\epsilon - (x_i - \mu_q)/\sqrt{v_q}$ to $(\epsilon - \hat\epsilon_i) + (\hat\epsilon_i - (x_i - \mu_q)/\sqrt{v_q})$ and recognizing that $\int \mathcal{N}(\epsilon;\hat\epsilon_i,\hat v_i)(\epsilon - \hat\epsilon_i)d\epsilon = 0$ while $(\hat\epsilon_i - (x_i - \mu_q)/\sqrt{v_q})$ is a constant with regards to $\epsilon$, we carry out the integration:
\begin{equation}
\widetilde G_{\mu_q} = -\sqrt{v_q}\sum_{i}\pi_i(\epsilon_i)\sqrt{\hat v_i}A_i\frac{\hat\epsilon_i - (x_i - \mu_q)/\sqrt{v_q}}{v_g} + \frac{(\mu_p - \mu_q)}{v_p}
\end{equation}
Next, we substitute $\hat\epsilon_i$ with its definition from equation (\ref{eq:hatepsiloni}) and rearrange:
\begin{equation}
\widetilde G_{\mu_q} = -\sum_{i}\pi_i(\epsilon_i)\sqrt{\hat v_i}A_i\left(\frac{v_q\left(1 - \pi_i(\epsilon_i)\hat v_i\right)}{v_g + v_q} - 1\right)\frac{x_i - \mu_q}{v_g} + \frac{(\mu_p - \mu_q)}{v_p}
\end{equation}
Further rearranging, we obtain the final expression for $\widetilde G_{\mu_q}$:
\begin{align}
&\widetilde G_{\mu_q} = \sum_{i}B_i\frac{x_i - \mu_q}{v_g} + \frac{(\mu_p - \mu_q)}{v_p} \\
&B_i = \pi_i(\epsilon_i)\sqrt{\hat v_i}A_i\frac{v_g + \pi_i(\epsilon_i)\hat v_iv_q}{v_g + v_q}
\end{align}

To compute the integral in equation (\ref{eq:elbogradv}) we substitute $q_0(\epsilon)\pi_i(\epsilon)$ with its approximation from equation (\ref{eq:q0piapprox}):
\begin{equation} \label{eq:elbogradv3}
G_{v_q} \approx \widetilde G_{v_q} = -\frac{1}{2}\left[\sum_{i}\pi_i(\epsilon_i)\sqrt{\hat v_i}A_i\int\mathcal{N}(\epsilon;\hat\epsilon_i,\hat v_i)\frac{\epsilon(\epsilon - (x_i - \mu_q)/\sqrt{v_q})}{v_g}d\epsilon - \frac{1}{v_q} + \frac{1}{v_p}\right]
\end{equation}
Modifying the term $\epsilon(\epsilon - (x_i - \mu_q)/\sqrt{v_q})$ to $((\epsilon - \hat\epsilon_i) + \hat\epsilon_i)((\epsilon - \hat\epsilon_i) + (\hat\epsilon_i - (x_i - \mu_q)/\sqrt{v_q}))$ we recognize that $\int \mathcal{N}\hat\epsilon_i(\epsilon;\hat\epsilon_i,\hat v_i)(\epsilon - \hat\epsilon_i)^2d\epsilon = \hat v_i$, $\int \mathcal{N}(\epsilon;\hat\epsilon_i,\hat v_i)(\epsilon - \hat\epsilon_i)d\epsilon = 0$, $\int \mathcal{N}(\epsilon;\hat\epsilon_i,\hat v_i)(\epsilon - \hat\epsilon_i)(\hat\epsilon_i - (x_i - \mu_q)/\sqrt{v_q})d\epsilon = 0$ and $\hat\epsilon_i(\hat\epsilon_i - (x_i - \mu_q)/\sqrt{v_q})$ is a constant with respect to $\epsilon$. Carrying out the integration in equation (\ref{eq:elbogradv3}) then results in the following expression:
\begin{equation} \label{eq:elbogradv4}
\widetilde G_{v_q} = -\frac{1}{2}\left[\sum_{i}\pi_i(\epsilon_i)\sqrt{\hat v_i}A_i\hat v_i/v_g + \sum_{i}\pi_i(\epsilon_i)\sqrt{\hat v_i}A_i\frac{\hat\epsilon_i(\hat\epsilon_i - (x_i - \mu_q)/\sqrt{v_q})}{v_g} - \frac{1}{v_q} + \frac{1}{v_p}\right]
\end{equation}
Next, we substitute $\hat\epsilon_i$ in $\hat\epsilon_i(\hat\epsilon_i - (x_i - \mu_q)/\sqrt{v_q})$ with its definition from equation (\ref{eq:hatepsiloni}) and rearrange:
\begin{align}
&\hat\epsilon_i\left(\hat\epsilon_i - \frac{x_i - \mu_q}{\sqrt{v_q}}\right) = \left(1 - \pi_i(\epsilon_i)\hat v_i\right)\frac{v_q\left(1 - \pi_i(\epsilon_i)\hat v_i\right) - v_g - v_q}{v_g + v_q}(x_i - \mu_q)^2\frac{1}{v_g + v_q} \\
&\hat\epsilon_i\left(\hat\epsilon_i - \frac{x_i - \mu_q}{\sqrt{v_q}}\right) = -\left(1 - \pi_i(\epsilon_i)\hat v_i\right)\frac{v_g + \pi_i(\epsilon_i)\hat v_iv_q}{v_g + v_q}(x_i - \mu_q)^2\frac{1}{v_g + v_q} \label{eq:partgvq}
\end{align}
Substituting equation (\ref{eq:partgvq}) back in equation (\ref{eq:elbogradv4}) we obtain the final expression for $\widetilde G_{v_q}$:
\begin{align}
&\widetilde G_{v_q} = \frac{1}{2}\left[-\sum_{i}C_i\frac{1}{v_g} + \sum_{i}D_i\frac{(x_i - \mu_q)^2}{v_g}\frac{1}{v_g + v_q} + \frac{1}{v_q} - \frac{1}{v_p}\right] \\
&C_i = \pi_i(\epsilon_i)\sqrt{\hat v_i}A_i\hat v_i \\
&D_i = (1 - \pi_i(\epsilon_i) \hat v_i)B_i
\end{align}

\end{document}